\documentclass[conference]{IEEEtran}

\usepackage{amsmath,amssymb,amsfonts}
\usepackage{graphicx}
\usepackage{textcomp}
\usepackage{xcolor}
\usepackage{booktabs}
\usepackage{multirow}
\usepackage{tikz}
\usetikzlibrary{shapes.geometric, arrows.meta, positioning, fit, calc, backgrounds, patterns, decorations.pathreplacing, decorations.markings, shadows}
\usepackage{url}
\usepackage{cite}

\begin{document}

\title{Radar-Camera BEV Multi-Task Learning with\\Cross-Task Attention Bridge for Joint\\3D Detection and Segmentation}

\author{
\IEEEauthorblockN{Ahmet İnanç\textsuperscript{1} \and Özgür Erkent\textsuperscript{1}}
}

\maketitle

\renewcommand{\thefootnote}{}
\footnotetext{\textsuperscript{1}Hacettepe University, Ankara, Turkey.\\
\{ahmetinanc, ozgurerkent\}@hacettepe.edu.tr}
\renewcommand{\thefootnote}{\arabic{footnote}}

\begin{abstract}
Bird's-eye-view (BEV) representations are the dominant paradigm for 3D perception in autonomous driving, providing a unified spatial canvas where detection and segmentation features are geometrically registered to the same physical coordinate system.
However, existing radar-camera fusion methods treat these tasks in isolation, missing the opportunity for cross-task feature sharing: object-level geometric cues from detection can sharpen segmentation, while dense road-layout context from segmentation can anchor detection.
We propose \textbf{CTAB} (Cross-Task Attention Bridge), a bidirectional module that exchanges features between detection and segmentation branches via multi-scale deformable attention in shared BEV space.
CTAB is integrated into a multi-task framework with an Instance Normalization-based segmentation decoder and learnable BEV upsampling to provide a more detailed BEV representation.
On nuScenes, CTAB improves segmentation on 7 classes over the joint multi-task baseline at essentially neutral detection.
On a 4-class subset (drivable area, pedestrian crossing, walkway, vehicle), our joint multi-task model achieves 51.0 mIoU-4 while simultaneously providing competitive 3D detection.
\end{abstract}


\section{Introduction}
\label{sec:intro}

Autonomous driving perception requires both 3D object detection and BEV map segmentation for safe planning.
While camera-only BEV methods~\cite{bevdet,bevdepth,beverse} have achieved strong results, radar sensors provide complementary depth and velocity information that is robust to adverse weather and lighting conditions~\cite{crn,rcbevdet}.
Radar-camera fusion in BEV space has emerged as an active research direction, with methods such as RCBEVDet~\cite{rcbevdet} and BEVCar~\cite{bevcar} achieving strong results for detection and segmentation, respectively.
However, all existing radar-camera methods address these tasks in isolation, missing the opportunity to share complementary information: detection features encode object-level geometry (centroids, sizes, orientations) that can sharpen segmentation boundaries, while segmentation features provide dense semantic context (road layout, drivable area) that can anchor detection.

Multi-task learning is essential for real-world deployment, where detection, segmentation, and planning must run simultaneously under strict latency budgets~\cite{milli2023multi}.
Running separate single-task models is prohibitively expensive: two independent radar-camera pipelines would double the backbone computation.
A shared BEV backbone amortizes feature extraction across tasks, and cross-task feature exchange can recover the performance gap that typically arises from multi-task optimization.
While multi-task BEV methods exist in camera-only~\cite{beverse,uniad} and LiDAR-camera settings~\cite{bevfusion,maskbev}, they typically treat task heads as independent consumers of shared features without explicit inter-task communication.
Cross-task interaction has proven effective in 2D dense prediction~\cite{padnet,mtinet,xu2022mtformer}; however, it remains unexplored in BEV 3D radar-camera perception, where detection and segmentation features are geometrically registered in the same physical coordinate system, making spatial cross-attention particularly natural.

We propose \textbf{CTAB} (Cross-Task Attention Bridge), a lightweight bidirectional module that enables explicit feature exchange between detection and segmentation in BEV space.
CTAB employs multi-scale deformable attention~\cite{zhudeformable} to let each task branch query spatially relevant features from the other.

Our contributions are:
\begin{enumerate}
    \item We present the first radar-camera BEV multi-task framework that jointly performs 3D object detection and BEV map segmentation using a shared radar-camera fused BEV representation.
    \item We introduce CTAB, a bidirectional cross-task attention module based on multi-scale deformable attention that enables explicit feature exchange between detection and segmentation branches.
    \item We demonstrate that CTAB improves both tasks over a strong multi-task baseline on nuScenes, with minimal computational overhead.
\end{enumerate}

\definecolor{camblue}{HTML}{4A90D9}
\definecolor{camlite}{HTML}{D6E6F8}
\definecolor{radarorg}{HTML}{E8923A}
\definecolor{radarlite}{HTML}{FAEACD}
\definecolor{fusiongrn}{HTML}{5CB85C}
\definecolor{fusionlite}{HTML}{D5F0D5}
\definecolor{ctabred}{HTML}{C0392B}
\definecolor{ctablite}{HTML}{F5D5D1}
\definecolor{segpurp}{HTML}{8E44AD}
\definecolor{seglite}{HTML}{E8D5F0}
\definecolor{detcyan}{HTML}{2980B9}
\definecolor{detlite}{HTML}{D4E9F7}
\definecolor{lossgold}{HTML}{F39C12}
\definecolor{losslite}{HTML}{FDF2D8}
\definecolor{cubetop}{HTML}{E8E8E8}
\definecolor{cubeside}{HTML}{C0C0C0}

\newcommand{\featcuboid}[8]{%
  \begin{scope}[shift={(#1,#2)}]
    \fill[#6, draw=black!60, line width=0.3pt] (0,0) rectangle (#3,#4);
    \fill[#6!40, draw=black!60, line width=0.3pt]
      (0,#4) -- ++(#5*0.35, #5*0.35) -- ++(#3,0) -- ++(-#5*0.35,-#5*0.35) -- cycle;
    \fill[#6!60, draw=black!60, line width=0.3pt]
      (#3,0) -- ++(#5*0.35,#5*0.35) -- ++(0,#4) -- ++(-#5*0.35,-#5*0.35) -- cycle;
    \node[font=\scriptsize, align=center] at (#3/2, #4/2+0.08) {#7};
    \node[font=\tiny, align=center, text=black!70] at (#3/2, #4/2-0.22) {#8};
  \end{scope}
}

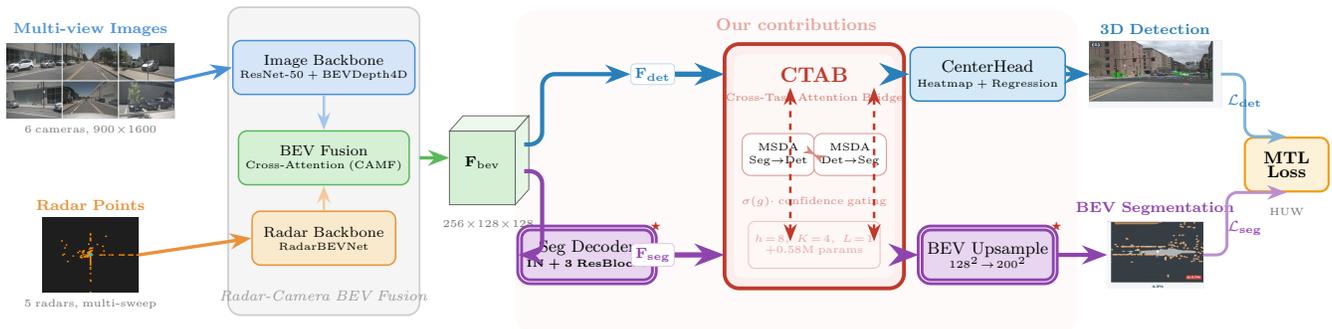
\begin{figure*}[!t]
\centering
\resizebox{0.97\textwidth}{!}{%
\begin{tikzpicture}[
    >=Stealth,
    yscale=1.12,
    arrow/.style={->, line width=1.2pt, black!65},
    modarrow/.style={->, line width=1.6pt},
    block/.style={draw, rounded corners=4pt, minimum height=0.9cm,
                  align=center, font=\footnotesize, line width=0.7pt,
                  general shadow={fill=black!8, shadow xshift=0.5pt,
                  shadow yshift=-0.5pt}},
    ours/.style={draw, rounded corners=5pt, minimum height=0.9cm,
                 align=center, font=\footnotesize, line width=1.4pt,
                 double, double distance=0.4pt,
                 general shadow={fill=black!10, shadow xshift=0.6pt,
                 shadow yshift=-0.6pt}},
    imgnode/.style={inner sep=0pt, outer sep=0pt,
                    draw=black!30, line width=0.5pt},
    lbl/.style={font=\tiny, text=black!55},
]

\node[imgnode] (camimg) at (0, 1.8)
  {\includegraphics[width=2.8cm, height=1.25cm]{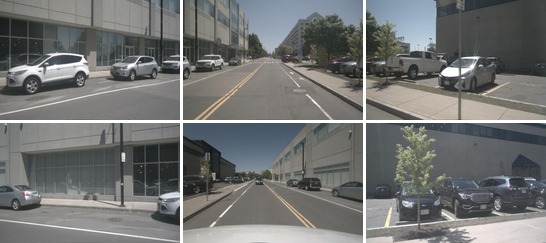}};
\node[font=\scriptsize\bfseries, text=camblue, anchor=south] at (camimg.north)
  {Multi-view Images};
\node[lbl, anchor=north] at (camimg.south) {6 cameras, $900\!\times\!1600$};

\node[imgnode] (radimg) at (0, -0.8)
  {\includegraphics[width=1.6cm, height=1.25cm]{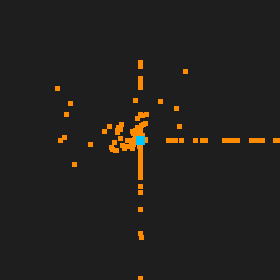}};
\node[font=\scriptsize\bfseries, text=radarorg, anchor=south] at (radimg.north)
  {Radar Points};
\node[lbl, anchor=north] at (radimg.south) {5 radars, multi-sweep};

\fill[black!4, rounded corners=6pt, draw=black!20, line width=0.8pt]
  (2.3, -1.7) rectangle (5.5, 2.9);

\node[block, fill=camlite, draw=camblue, minimum width=2.4cm, font=\scriptsize]
  (imgbb) at (3.9, 2.0) {Image Backbone\\[-2pt]{\tiny ResNet-50 + BEVDepth4D}};
\node[block, fill=fusionlite, draw=fusiongrn, minimum width=2.4cm, font=\scriptsize]
  (fuse) at (3.9, 0.65) {BEV Fusion\\[-2pt]{\tiny Cross-Attention (CAMF)}};
\node[block, fill=radarlite, draw=radarorg, minimum width=2.4cm, font=\scriptsize]
  (radbb) at (3.9, -0.55) {Radar Backbone\\[-2pt]{\tiny RadarBEVNet}};

\node[font=\scriptsize, text=black!40, anchor=south] at (3.9, -1.6)
  {\textit{Radar-Camera BEV Fusion}};

\featcuboid{6.0}{-0.05}{1.1}{1.1}{0.65}{fusionlite}{$\mathbf{F}_{\text{bev}}$}{}
\node[font=\tiny, text=black!55] at (6.65, -0.35)
  {$256\!\times\!128\!\times\!128$};

\node[ours, fill=seglite, draw=segpurp, minimum width=2.2cm]
  (segdec) at (8.3, -0.8) {Seg Decoder\\[-2pt]{\tiny\bfseries IN + 3 ResBlocks}};

\begin{scope}[shift={(11.3, 0.35)}]
  \fill[ctablite!40, rounded corners=7pt, draw=ctabred, line width=1.8pt]
    (-0.7, -1.65) rectangle (1.8, 2.0);
  \fill[ctablite!20, rounded corners=5pt]
    (-0.55, -1.5) rectangle (1.65, 1.85);

  \node[font=\normalsize\bfseries, text=ctabred] at (0.5, 1.55) {CTAB};
  \node[font=\tiny, text=ctabred!70, align=center] at (0.5, 1.2)
    {Cross-Task\\[-1pt]Attention Bridge};

  \node[block, fill=white, draw=ctabred!50, minimum width=0.7cm,
        minimum height=0.7cm, font=\tiny, line width=0.5pt]
    (msda_s2d) at (-0.1, 0.35) {MSDA\\Seg$\to$Det};
  \node[block, fill=white, draw=ctabred!50, minimum width=0.7cm,
        minimum height=0.7cm, font=\tiny, line width=0.5pt]
    (msda_d2s) at (1.1, 0.35) {MSDA\\Det$\to$Seg};

  \draw[ctabred!60, ->, line width=0.6pt, bend left=25]
    (msda_s2d.east) to (msda_d2s.west);
  \draw[ctabred!60, ->, line width=0.6pt, bend left=25]
    (msda_d2s.west) to (msda_s2d.east);

  \node[font=\tiny, text=ctabred!60, align=center] at (0.5, -0.35)
    {$\sigma(g)\cdot$ confidence gating};

  \draw[ctabred!30, line width=0.5pt, rounded corners=3pt]
    (-0.6, -1.35) rectangle (1.6, -0.65);
  \node[font=\tiny, text=ctabred!45, align=center] at (0.5, -1.0)
    {$h\!=\!8,\;K\!=\!4,\;L\!=\!1$\\[-1pt]+0.58M params};
\end{scope}

\node[block, fill=detlite, draw=detcyan, minimum width=1.8cm]
  (centerhead) at (15.0, 1.9) {CenterHead\\[-2pt]{\tiny Heatmap + Regression}};
\node[ours, fill=seglite, draw=segpurp, minimum width=1.8cm]
  (bevup) at (15.0, -0.8) {BEV Upsample\\[-2pt]{\tiny\bfseries $128^2\!\to\!200^2$}};

\node[imgnode] (detout) at (17.8, 1.9)
  {\includegraphics[width=1.8cm, height=1.1cm]{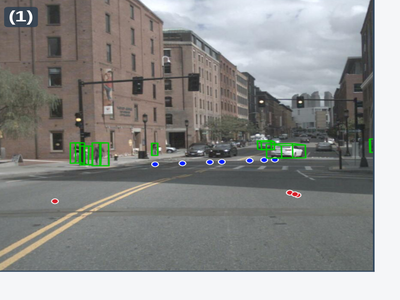}};
\node[font=\scriptsize\bfseries, text=detcyan, anchor=south] at (detout.north)
  {3D Detection};

\node[imgnode] (segout) at (17.8, -0.8)
  {\includegraphics[width=1.6cm, height=1.1cm]{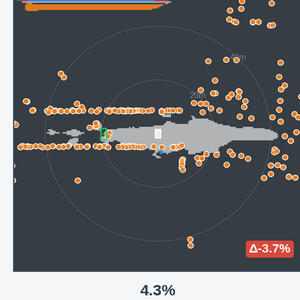}};
\node[font=\scriptsize\bfseries, text=segpurp, anchor=south] at (segout.north)
  {BEV Segmentation};

\node[block, fill=losslite, draw=lossgold, minimum width=1.4cm, minimum height=0.9cm,
      line width=1pt, font=\footnotesize\bfseries]
  (mtlloss) at (20.0, 0.55) {MTL\\[-2pt]Loss};
\node[lbl] at (20.7, -0.05) {HUW};

\draw[modarrow, camblue] (camimg.east) -- (imgbb.west);
\draw[modarrow, radarorg] (radimg.east) -- (radbb.west);
\draw[arrow, camblue!50] (imgbb.south) -- (fuse.north);
\draw[arrow, radarorg!50] (radbb.north) -- (fuse.south);

\draw[modarrow, fusiongrn] (5.5, 0.65) -- (6.0, 0.65);

\draw[->, line width=2pt, detcyan, rounded corners=6pt]
  (7.25, 0.85) -- (7.55, 0.85) |- (9.7, 1.9);
\draw[->, line width=2pt, segpurp, rounded corners=6pt]
  (7.25, 0.45) -| (segdec.north);

\draw[->, line width=2.5pt, detcyan] (9.7, 1.9) -- (10.6, 1.9);
\draw[->, line width=2.5pt, detcyan] (13.1, 1.9) -- (centerhead.west);

\draw[->, line width=2.5pt, segpurp] (10.0, -0.8) -- (10.6, -0.8);
\draw[->, line width=2.5pt, segpurp] (13.1, -0.8) -- (bevup.west);

\begin{scope}[on background layer]
  \draw[dashed, line width=1.2pt, detcyan!70] (10.6, 1.9) -- (13.1, 1.9);
  \draw[dashed, line width=1.2pt, segpurp!70] (10.6, -0.8) -- (13.1, -0.8);
\end{scope}

\draw[ctabred, <->, line width=1.2pt, dashed]
  (11.4, 1.7) -- (11.4, -0.6);
\draw[ctabred, <->, line width=1.2pt, dashed]
  (12.8, 1.7) -- (12.8, -0.6);

\draw[->, line width=2pt, detcyan] (centerhead.east) -- (detout.west);
\draw[->, line width=2pt, segpurp] (bevup.east) -- (segout.west);

\node[font=\scriptsize\bfseries, text=detcyan, fill=white,
      inner sep=2pt, rounded corners=2pt, draw=detcyan!30, line width=0.3pt]
  (fdet) at (9.4, 1.9) {$\mathbf{F}_{\text{det}}$};
\node[font=\scriptsize\bfseries, text=segpurp, fill=white,
      inner sep=2pt, rounded corners=2pt, draw=segpurp!30, line width=0.3pt]
  (fseg) at (9.7, -0.8) {$\mathbf{F}_{\text{seg}}$};

\draw[->, line width=1.4pt, detcyan!60, rounded corners=5pt]
  (detout.east) -| (mtlloss.north);
\draw[->, line width=1.4pt, segpurp!60, rounded corners=5pt]
  (segout.east) -| (mtlloss.south);

\node[font=\scriptsize\bfseries, text=detcyan] at (19.6, 1.4)
  {$\mathcal{L}_{\text{det}}$};
\node[font=\scriptsize\bfseries, text=segpurp] at (19.6, -0.4)
  {$\mathcal{L}_{\text{seg}}$};

\begin{scope}[on background layer]
  \fill[ctablite!12, rounded corners=10pt]
    (7.1, -2.0) rectangle (16.5, 2.85);
\end{scope}
\node[font=\footnotesize\bfseries, text=ctabred!45,
      fill=ctablite!12, inner sep=2pt, rounded corners=3pt]
  at (11.8, 2.65) {Our contributions};

\node[font=\tiny, text=ctabred] at (segdec.north east) {$\bigstar$};
\node[font=\tiny, text=ctabred] at (bevup.north east) {$\bigstar$};

\end{tikzpicture}%
}
\caption{\textbf{Overall architecture.}
Multi-view camera images and radar point clouds are processed by a backbone that combines an image backbone with a radar backbone to create a BEV fusion similar to RCBEVDet, yielding a shared BEV feature $\mathbf{F}_{\text{bev}} \in \mathbb{R}^{256 \times 128 \times 128}$.
The detection path operates directly on $\mathbf{F}_{\text{bev}}$ in BEV coordinates; the segmentation decoder ($\bigstar$) extracts $\mathbf{F}_{\text{seg}}$ with Instance Normalization.
\textbf{CTAB} ($\bigstar$) exchanges features bidirectionally via gated multi-scale deformable attention.
CenterHead predicts 3D boxes from BEV features; BEV Upsampling ($\bigstar$) refines segmentation to $200 \times 200$.
HUW balances both objectives with learned task uncertainty weights.
}
\label{fig:architecture}
\end{figure*}

\section{Related Work}
\label{sec:related}

\subsection{Radar-Camera BEV Fusion}

Radar-camera fusion combines dense visual information with radar's direct depth and velocity measurements, offering a cost-effective alternative to LiDAR.
BEV-based fusion has become dominant, projecting both modalities into a shared spatial representation.
For detection, CRN~\cite{crn} introduced radar-assisted view transformation, and RCBEVDet~\cite{rcbevdet} proposed dual-stream radar encoding with cross-attention fusion.
Subsequent methods have pushed performance further: HyDRa~\cite{wolters2025unleashing} combines image-level and BEV-level fusion, CRT-Fusion~\cite{kim2024crt} adds temporal aggregation, RCBEVDet++~\cite{lin2024rcbevdet++} enhances radar encoding with separate task heads, and RaCFormer~\cite{chu2025racformer} uses query-based fusion.

For segmentation, Simple-BEV~\cite{harley2023simple} showed that adding radar improves vehicle IoU by +8.3. BEVGuide~\cite{Man_2023_CVPR} uses a BEV-guided multi-sensor attention module.
BEVCar~\cite{bevcar} adopts a lifting mechanism that projects image features into the BEV space using deformable attention. RESAR-BEV~\cite{zeng2026resar} introduces a progressive residual autoregressive approach with geometry-guided radar feature querying.

Critically, all these methods address either detection or segmentation in isolation.
RCBEVDet++~\cite{lin2024rcbevdet++} produces both outputs but with independent heads and no cross-task interaction.
\textbf{No prior radar-camera method jointly optimizes detection and segmentation with explicit inter-task feature exchange.}

\subsection{Multi-Task BEV Perception}
BEV representations naturally support multi-task learning by providing a unified spatial canvas.
Camera-only methods such as BEVerse~\cite{beverse} and HENet~\cite{xia2024henet} perform joint detection and segmentation, while end-to-end planners like UniAD~\cite{uniad} chain multiple perception tasks with planning but do not optimize inter-task feature exchange.
In the LiDAR-camera domain, BEVFusion~\cite{bevfusion} uses independent task heads, FULLER~\cite{huang2023fuller} addresses negative transfer via gradient calibration, MetaBEV~\cite{ge2023metabev} tackles sensor-failure robustness, while M\textsuperscript{3}Net~\cite{chen2025m3net} and MaskBEV~\cite{maskbev} introduce cross-task mechanisms in a LiDAR-centric setting.

Cross-task interaction in 2D dense prediction has been explored via mutual distillation~\cite{padnet}, multi-scale interactions~\cite{mtinet}, transformer cross-attention~\cite{xu2022mtformer}, and task-specific prompting~\cite{ye2023taskprompter}.
These methods demonstrate that explicit bidirectional feature exchange outperforms naive parameter sharing; however, such mechanisms remain unexplored in BEV 3D perception with \emph{radar}-camera fusion, where detection and segmentation features are geometrically aligned and share complementary radar-camera signals.
\textbf{CTAB bridges this gap} by adapting deformable cross-attention to the BEV coordinate system of a radar-camera framework.

Multi-task loss balancing is a fundamental challenge.
Kendall~et~al.~\cite{kendall2018multi} proposed homoscedastic uncertainty weighting, learning per-task noise parameters to automatically balance loss magnitudes.
GradNorm~\cite{chen2018gradnorm} dynamically normalizes gradient magnitudes across tasks, while PCGrad~\cite{yu2020gradient} projects conflicting gradients to reduce negative transfer.
We adopt homoscedastic uncertainty weighting (HUW)~\cite{kendall2018multi} for its simplicity and ease of integration into multi-task BEV frameworks.

\section{Method}
\label{sec:method}

\subsection{Overview}

Our framework (Fig.~\ref{fig:architecture}) extends the RCBEVDet~\cite{rcbevdet} detector with a BEV segmentation branch and the proposed CTAB module.
The overall pipeline proceeds as follows.

\textbf{Shared BEV backbone.}
The camera branch uses a pipeline similar to BEVDepth~\cite{bevdepth}: a ResNet-50 backbone extracts multi-view image features, which are projected to BEV via explicit depth estimation and voxel pooling, with temporal stereo from adjacent frames.
The radar branch processes multi-sweep radar points through RadarBEVNet with dual-stream encoding and cross-modality Injection-and-Extraction modules.
Camera and radar BEV features are fused via cross-attention into a shared feature map $\mathbf{F}_{\text{bev}} \in \mathbb{R}^{B \times 256 \times 128 \times 128}$, covering $102.4\,\text{m} \times 102.4\,\text{m}$ at $0.8\,\text{m}$ resolution.

\textbf{Task branches.}
The detection branch consumes $\mathbf{F}_{\text{bev}}$ directly at $128 \times 128$ via CenterHead~\cite{yin2021center}.
The segmentation branch first extracts intermediate features through a dedicated decoder, then upsamples to $200 \times 200$ ($0.5\,\text{m/cell}$) via the BEV Upsampling module (Section~\ref{sec:bev_upsample}).
When enabled, CTAB facilitates bidirectional feature exchange between the two branches before upsampling and final predictions.

\subsection{BEV Segmentation Decoder}

The segmentation decoder transforms $\mathbf{F}_{\text{bev}}$ into per-pixel class predictions for $C$ BEV map classes.
The design is class-agnostic and supports any BEV label set; the specific classes we evaluate, together with how vehicle masks are generated from 3D bounding box annotations, are described in Section~\ref{sec:experiments}.

The decoder architecture consists of an input projection (Conv $3\!\times\!3$, $256 \to 128$, IN, ReLU) followed by three residual blocks (each with two Conv $3\!\times\!3$ layers, Instance Normalization, and a skip connection) and a prediction head (Conv $3\!\times\!3$, IN, ReLU, then Conv $1\!\times\!1$, $128 \to C$).

\textbf{Normalization choice.}
We use Instance Normalization (IN) instead of Batch Normalization (BN) throughout all modules trained from scratch (decoder, CTAB, upsampling).
Since these modules initialize with random weights alongside a pretrained detection backbone, BN's running statistics are unreliable in early epochs and diverge between training and evaluation modes---particularly problematic in multi-view BEV settings where effective batch diversity is limited by the multi-camera structure within each sample.
IN normalizes each sample independently, eliminating this train-eval discrepancy.
For CTAB's projection layers, we use Group Normalization (GN)~\cite{wu2018group} for the same reason; GN is preferred over IN in attention modules as it allows channel grouping.

The intermediate feature after the third residual block, $\mathbf{F}_{\text{seg}} \in \mathbb{R}^{B \times 128 \times H \times W}$, serves as the segmentation input to CTAB, providing rich semantic features before the final classification layer.

\subsection{BEV Upsampling}
\label{sec:bev_upsample}

The pretrained fusion backbone produces BEV features at $128 \times 128$ resolution ($0.8\,\text{m/cell}$), which is insufficient for thin map structures such as lane dividers (${\sim}15\,\text{cm}$ width) and stop lines (${\sim}30\,\text{cm}$).
At $0.8\,\text{m}$ resolution, these structures occupy sub-pixel widths, making their segmentation inherently noisy.
We introduce a lightweight, task-specific BEV Upsampling module that operates only on the segmentation branch.
This makes it compatible with other pretrained fusion backbones without modification.

The module upsamples the intermediate segmentation feature $\mathbf{F}_{\text{seg}} \in \mathbb{R}^{B \times 128 \times 128 \times 128}$ to $200 \times 200$ ($0.5\,\text{m/cell}$) using bilinear interpolation followed by a learnable residual refinement block:
\begin{gather}
    \mathbf{F}_{\text{seg}}^{\uparrow} = \mathbf{U} + \text{IN}(\text{Conv}_{3\times3}(\text{ReLU}(\text{IN}(\text{Conv}_{3\times3}(\mathbf{U}))))) \notag\\
    \text{where}\;\; \mathbf{U} = \text{Upsample}(\mathbf{F}_{\text{seg}})
\end{gather}
The residual path refines the bilinearly upsampled features, recovering high-frequency spatial detail that interpolation alone cannot provide.
This adds only 0.30M parameters while enabling the segmentation branch to predict at $200 \times 200$ resolution, matching the evaluation grid used by other methods such as BEVCar~\cite{bevcar}.
The detection branch continues to operate at the native $128 \times 128$ resolution, preserving the pretrained CenterHead's spatial assumptions.

\subsection{Cross-Task Attention Bridge (CTAB)}
\label{sec:ctab}

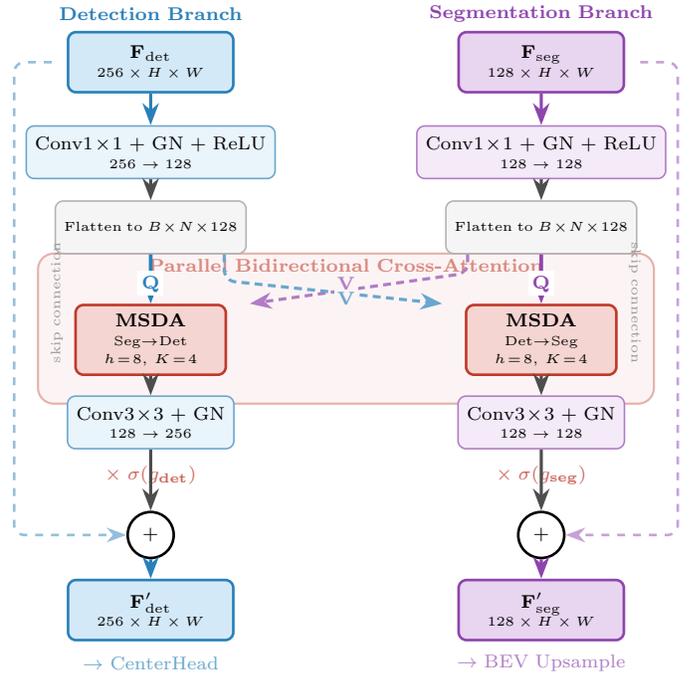
\begin{figure}[t]
\centering
\resizebox{\columnwidth}{!}{%
\begin{tikzpicture}[
    >=Stealth,
    arr/.style={->, line width=1.2pt, black!70},
    blk/.style={draw, rounded corners=3pt, minimum height=0.7cm,
                align=center, font=\scriptsize, line width=0.6pt},
    skip/.style={->, dashed, black!30, line width=1pt},
    gatelbl/.style={font=\scriptsize\bfseries, text=ctabred!80,
                    fill=white, inner sep=2pt, rounded corners=2pt},
]

\def\colL{0.0}    
\def\colR{5.2}    
\def\midX{2.6}    

\node[blk, fill=detlite, draw=detcyan, minimum width=2.2cm,
      minimum height=0.8cm, line width=1pt]
  (fdet) at (\colL, 5.5) {$\mathbf{F}_{\text{det}}$\\[-1pt]{\tiny$256 \times H \times W$}};
\node[blk, fill=seglite, draw=segpurp, minimum width=2.2cm,
      minimum height=0.8cm, line width=1pt]
  (fseg) at (\colR, 5.5) {$\mathbf{F}_{\text{seg}}$\\[-1pt]{\tiny$128 \times H \times W$}};

\node[blk, fill=detlite!50, draw=detcyan!70, minimum width=2.2cm]
  (dproj) at (\colL, 4.3) {Conv$1\!\times\!1$ + GN + ReLU\\[-1pt]{\tiny$256 \to 128$}};
\node[blk, fill=seglite!50, draw=segpurp!70, minimum width=2.2cm]
  (sproj) at (\colR, 4.3) {Conv$1\!\times\!1$ + GN + ReLU\\[-1pt]{\tiny$128 \to 128$}};

\draw[arr, detcyan] (fdet) -- (dproj);
\draw[arr, segpurp] (fseg) -- (sproj);

\node[blk, fill=black!4, draw=black!30, minimum width=2.2cm, font=\tiny]
  (dflt) at (\colL, 3.3) {Flatten to $B\!\times\!N\!\times\!128$};
\node[blk, fill=black!4, draw=black!30, minimum width=2.2cm, font=\tiny]
  (sflt) at (\colR, 3.3) {Flatten to $B\!\times\!N\!\times\!128$};

\draw[arr] (dproj) -- (dflt);
\draw[arr] (sproj) -- (sflt);

\begin{scope}[on background layer]
  \fill[ctablite!25, rounded corners=6pt, draw=ctabred!40, line width=0.8pt]
    (-1.5, 0.95) rectangle (6.7, 2.95);
\end{scope}
\node[font=\scriptsize\bfseries, text=ctabred!60] at (\midX, 2.8)
  {Parallel Bidirectional Cross-Attention};

\node[blk, fill=ctablite, draw=ctabred, minimum width=2.0cm,
      minimum height=0.9cm, line width=1pt]
  (msda_s2d) at (\colL, 1.8) {\textbf{MSDA}\\[-1pt]{\tiny Seg$\to$Det}\\[-1pt]{\tiny$h\!=\!8,\,K\!=\!4$}};

\node[blk, fill=ctablite, draw=ctabred, minimum width=2.0cm,
      minimum height=0.9cm, line width=1pt]
  (msda_d2s) at (\colR, 1.8) {\textbf{MSDA}\\[-1pt]{\tiny Det$\to$Seg}\\[-1pt]{\tiny$h\!=\!8,\,K\!=\!4$}};


\draw[arr, detcyan, line width=1.4pt]
  (dflt.south) -- (msda_s2d.north);
\node[font=\scriptsize\bfseries, text=detcyan, fill=white, inner sep=1.5pt]
  at (\colL, 2.55) {Q};

\draw[arr, segpurp, line width=1.4pt]
  (sflt.south) -- (msda_d2s.north);
\node[font=\scriptsize\bfseries, text=segpurp, fill=white, inner sep=1.5pt]
  at (\colR, 2.55) {Q};

\draw[->, line width=1.2pt, segpurp!70, densely dashed, rounded corners=4pt]
  ([xshift=0.3cm]sflt.south west) -- ++(0, -0.25)
  -- ([xshift=0.3cm]msda_s2d.north east);
\node[font=\scriptsize\bfseries, text=segpurp!70, fill=ctablite!25, inner sep=1pt]
  at (\midX, 2.55) {V};

\draw[->, line width=1.2pt, detcyan!70, densely dashed, rounded corners=4pt]
  ([xshift=-0.3cm]dflt.south east) -- ++(0, -0.35)
  -- ([xshift=-0.3cm]msda_d2s.north west);
\node[font=\scriptsize\bfseries, text=detcyan!70, fill=ctablite!25, inner sep=1pt]
  at (\midX, 2.35) {V};

\node[blk, fill=detlite!50, draw=detcyan!70, minimum width=2.2cm]
  (dout) at (\colL, 0.7) {Conv$3\!\times\!3$ + GN\\[-1pt]{\tiny$128 \to 256$}};
\node[blk, fill=seglite!50, draw=segpurp!70, minimum width=2.2cm]
  (sout) at (\colR, 0.7) {Conv$3\!\times\!3$ + GN\\[-1pt]{\tiny$128 \to 128$}};

\draw[arr] (msda_s2d) -- (dout);
\draw[arr] (msda_d2s) -- (sout);

\node[draw, circle, minimum size=0.45cm, line width=1pt, fill=white,
      font=\scriptsize\bfseries] (addL) at (\colL, -0.8) {$+$};
\node[draw, circle, minimum size=0.45cm, line width=1pt, fill=white,
      font=\scriptsize\bfseries] (addR) at (\colR, -0.8) {$+$};

\draw[arr] (dout.south) -- (addL.north);
\draw[arr] (sout.south) -- (addR.north);

\node[gatelbl] at (-0.9, 0.0) {$\times\;\sigma(g_{\text{det}})$};
\node[gatelbl] at (6.1, 0.0) {$\times\;\sigma(g_{\text{seg}})$};

\draw[skip, detcyan!50, rounded corners=4pt]
  ([xshift=-0.2cm]fdet.west) -- ++(-0.5, 0) |- (addL.west);
\draw[skip, segpurp!50, rounded corners=4pt]
  ([xshift=0.2cm]fseg.east) -- ++(0.5, 0) |- (addR.east);

\node[font=\tiny, text=black!40, rotate=90] at (-1.65, 2.3) {skip connection};
\node[font=\tiny, text=black!40, rotate=-90] at (6.85, 2.3) {skip connection};

\node[blk, fill=detlite, draw=detcyan, minimum width=2.2cm,
      minimum height=0.8cm, line width=1pt]
  (fdet_out) at (\colL, -1.8) {$\mathbf{F}'_{\text{det}}$\\[-1pt]{\tiny$256 \times H \times W$}};
\node[blk, fill=seglite, draw=segpurp, minimum width=2.2cm,
      minimum height=0.8cm, line width=1pt]
  (fseg_out) at (\colR, -1.8) {$\mathbf{F}'_{\text{seg}}$\\[-1pt]{\tiny$128 \times H \times W$}};

\draw[arr, detcyan] (addL) -- (fdet_out);
\draw[arr, segpurp] (addR) -- (fseg_out);

\node[font=\scriptsize, text=detcyan!70] at (\colL, -2.5) {$\to$ CenterHead};
\node[font=\scriptsize, text=segpurp!70] at (\colR, -2.5) {$\to$ BEV Upsample};

\node[font=\scriptsize\bfseries, text=detcyan] at (\colL, 6.15)
  {Detection Branch};
\node[font=\scriptsize\bfseries, text=segpurp] at (\colR, 6.15)
  {Segmentation Branch};

\end{tikzpicture}%
}
\caption{\textbf{Detailed architecture of the CTAB module.}
Detection and segmentation features are projected to a shared $d\!=\!128$ space and flattened.
Two parallel MSDA blocks perform bidirectional cross-attention: in Seg$\to$Det, detection features serve as queries attending to segmentation values; in Det$\to$Seg, the roles are reversed.
Output convolutions project back to original dimensions.
Confidence gates $\sigma(g)$, initialized at ${\approx}0.12$, scale the cross-attention output before residual addition, preventing destabilization of pretrained features.
}
\label{fig:ctab}
\end{figure}

CTAB enables bidirectional feature exchange between detection and segmentation branches using multi-scale deformable attention~\cite{zhudeformable} (Fig.~\ref{fig:ctab}).

\textbf{Design motivation.}
In our BEV setting, detection and segmentation features are \textit{geometrically registered} to the same physical coordinate system, making spatial cross-attention particularly well-suited: a query at BEV position $(x, y)$ can directly sample task-complementary information from relevant locations.
We choose deformable attention over standard cross-attention for its linear scaling ($\mathcal{O}(NK)$ vs.\ $\mathcal{O}(N^2)$), critical for BEV grids where $N = 128^2 = 16{,}384$.
Learned sampling offsets allow each query to adaptively attend to non-local regions, capturing spatial relationships between objects and their road context.

\subsubsection{Input Projection}
The detection feature $\mathbf{F}_{\text{det}} \in \mathbb{R}^{B \times 256 \times H \times W}$ and segmentation feature $\mathbf{F}_{\text{seg}} \in \mathbb{R}^{B \times 128 \times H \times W}$ are projected to a shared dimension $d = 128$:
\begin{align}
    \hat{\mathbf{F}}_{\text{det}} &= \text{ReLU}(\text{GN}(\text{Conv}_{1\times 1}(\mathbf{F}_{\text{det}}))) \\
    \hat{\mathbf{F}}_{\text{seg}} &= \text{ReLU}(\text{GN}(\text{Conv}_{1\times 1}(\mathbf{F}_{\text{seg}})))
\end{align}
Note that the detection projection reduces from 256 to 128 channels, while the segmentation projection retains 128 channels; the shared hidden dimension enables direct cross-attention between the two branches.
Both features are then flattened to sequences $\hat{\mathbf{F}} \in \mathbb{R}^{B \times N \times d}$, where $N = H \times W$.

\subsubsection{Bidirectional Deformable Cross-Attention}
The core of CTAB is Multi-Scale Deformable Attention (MSDA)~\cite{zhudeformable}: each query learns $K$ spatial offsets and samples values at those positions via bilinear interpolation (instead of attending to all $N$ locations), across $h$ parallel heads operating on $d/h$-dim sub-channels, with $L$ denoting the number of feature scales aggregated per query.
We use $h\!=\!8$ heads (16 channels each at $d\!=\!128$, following DETR-style defaults), $K\!=\!4$ sampling points (sufficient to cover typical object footprints in BEV at low cost), and $L\!=\!1$ since the shared fused BEV is a single-resolution feature map.
CTAB applies two parallel MSDA operations:
\begin{align}
    \mathbf{A}_{\text{d2s}} &= \text{MSDA}(\mathbf{Q} = \hat{\mathbf{F}}_{\text{seg}},\; \mathbf{V} = \hat{\mathbf{F}}_{\text{det}},\; \mathbf{p}_{\text{ref}}) \\
    \mathbf{A}_{\text{s2d}} &= \text{MSDA}(\mathbf{Q} = \hat{\mathbf{F}}_{\text{det}},\; \mathbf{V} = \hat{\mathbf{F}}_{\text{seg}},\; \mathbf{p}_{\text{ref}})
\end{align}
In Det$\to$Seg, segmentation queries attend to detection features to incorporate object-level geometric information (centroids, sizes, orientations).
In Seg$\to$Det, detection queries attend to segmentation features to incorporate dense road layout and drivable area context.
Both operations run in parallel, sharing reference points $\mathbf{p}_{\text{ref}} \in [0, 1]^{N \times 2}$ generated as a uniform grid over the BEV spatial dimensions.
We explicitly disable MSDA's internal residual connection by passing a zero identity tensor, ensuring that only the learned cross-attention signal passes through to the output projection.

\subsubsection{Gated Output Projection}
The attended features are projected back to original channel dimensions via confidence-gated residual connections:
\begin{align}
    \mathbf{F}_{\text{det}}^{\prime} &= \sigma(g_{\text{det}}) \cdot \text{GN}(\text{Conv}_{3\times 3}(\mathbf{A}_{\text{s2d}})) + \mathbf{F}_{\text{det}} \\
    \mathbf{F}_{\text{seg}}^{\prime} &= \sigma(g_{\text{seg}}) \cdot \text{GN}(\text{Conv}_{3\times 3}(\mathbf{A}_{\text{d2s}})) + \mathbf{F}_{\text{seg}}
\end{align}
where $\sigma(\cdot)$ denotes the logistic sigmoid function that maps each scalar gate to the range $(0,1)$.
The scalar gates $g_{\text{det}}, g_{\text{seg}}$ are learnable parameters initialized to $-2.0$, yielding $\sigma(-2.0) \approx 0.12$ at the start of training.
This initialization prevents the untrained cross-attention signal from destabilizing the pretrained detection features and the newly initialized segmentation features.
As the attention weights converge during training, the network learns to open the gates, with the rate of increase reflecting how beneficial each cross-task direction is.
No activation function is applied before the gated residual addition, allowing the cross-attention output to both amplify and suppress features through positive and negative corrections.
The enhanced $\mathbf{F}_{\text{det}}^{\prime}$ proceeds to CenterHead for 3D detection, while $\mathbf{F}_{\text{seg}}^{\prime}$ proceeds to the BEV Upsampling module and segmentation prediction head.

\subsection{Multi-Task Loss}

We balance tasks via homoscedastic uncertainty weighting (HUW)~\cite{kendall2018multi}:
\begin{equation}
    \mathcal{L}_{\text{total}} = \frac{1}{2\sigma_{\text{det}}^2} \mathcal{L}_{\text{det}} + \log \sigma_{\text{det}} + \frac{1}{2\sigma_{\text{seg}}^2} \mathcal{L}_{\text{seg}} + \log \sigma_{\text{seg}}
    \label{eq:kendall}
\end{equation}
where $\sigma_{\text{det}}, \sigma_{\text{seg}}$ are learnable parameters ($\log \sigma$ clamped at 1.5).
$\mathcal{L}_{\text{det}}$ uses CenterHead~\cite{yin2021center} losses (heatmap focal, regression L1, depth).
$\mathcal{L}_{\text{seg}}$ combines sigmoid focal loss ($\alpha\!=\!0.25$, $\gamma\!=\!3$, following BEVCar~\cite{bevcar}) and dice loss, each weighted by 20.
GT masks are generated at $200 \times 200$ to match the upsampled output.

\section{Experiments}
\label{sec:experiments}

\subsection{Dataset and Metrics}

We evaluate on the nuScenes dataset~\cite{caesar2020nuscenes}, which provides 1000 driving scenes captured with 6 cameras and 5 radar sensors in Boston and Singapore.
We use the standard train/val split (28,130 / 6,019 samples).
For 3D detection, we report nuScenes Detection Score (NDS) and mean Average Precision (mAP).
For BEV segmentation, we report mean Intersection-over-Union (mIoU) across 7 map classes (mIoU-7): drivable area, carpark area, pedestrian crossing, walkway, stop line, divider, and vehicle.
We also report a 4-class mIoU (mIoU-4: drivable area, pedestrian crossing, walkway, vehicle).
Note that prior segmentation methods use varying class definitions: BEVCar~\cite{bevcar} reports a 2-class mean (vehicle, drivable), BEVGuide~\cite{Man_2023_CVPR} and RCBEVDet++~\cite{lin2024rcbevdet++} use 3 classes (vehicle, drivable, lane), and RESAR-BEV~\cite{zeng2026resar} uses 7 classes with a different set (no carpark, separate road/lane dividers), precluding direct mIoU comparison across methods.

\subsection{Implementation Details}

Our framework is implemented in PyTorch with mmdetection3d~\cite{contributors2020mmdetection3d}.
We use ResNet-50 as the image backbone, initialized with ImageNet pretrained weights; the detection backbone is further initialized from a pretrained RCBEVDet~\cite{rcbevdet} checkpoint trained on the same BEV grid.
The shared BEV grid covers $[-51.2, 51.2]\,\text{m}$ at $128 \times 128$ ($0.8\,\text{m/cell}$); the segmentation branch upsamples to $200 \times 200$ ($0.5\,\text{m/cell}$).
Training uses AdamW with cosine annealing (peak lr $4 \times 10^{-4}$, linear warmup over 500 iterations) for 10 epochs, batch size 16, and exponential moving average (EMA) on an NVIDIA A6000 GPU.

Data augmentation follows RCBEVDet~\cite{rcbevdet}: random flip, rotation ($\pm 22.5^\circ$), and scaling ($0.95$--$1.05\times$) in BEV space.
These BEV data augmentation (BDA) transforms are applied to both 3D frustum points and detection GT boxes.
We additionally apply the same affine transform to pre-computed segmentation GT masks via nearest-neighbor interpolation, ensuring pixel-level alignment between augmented BEV features and segmentation targets.

The segmentation decoder has 1.33M parameters (3 residual blocks with IN).
The BEV Upsampling module adds 0.30M parameters.
The CTAB module ($d=128$, $h=8$, $K=4$) adds 0.58M parameters.
Total overhead: +2.21M for the full MTL+CTAB framework over detection-only.
HUW balances the multi-task loss with $\log \sigma$ clamped at 1.5 to ensure numerical stability while permitting sufficient dynamic range.

\subsection{Main Results}

\begin{table}[t]
\centering
\caption{Radar-camera (R+C) methods on nuScenes val. $*$: single-task model. $\ddagger$: cross-task interaction. mIoU-7: our 7-class evaluation. mIoU-4: our 4-class subset (drivable, ped.\ crossing, walkway, vehicle). Seg mIoU$^*$ from other works uses their own class definitions: $^a$2-class (vehicle, drivable), $^b$3-class (vehicle, drivable, lane), $^c$7-class (different class set).}
\label{tab:comparison}
\renewcommand{\arraystretch}{1.15}
\resizebox{\columnwidth}{!}{%
\begin{tabular}{@{}l c c cc cc@{}}
\toprule
\textbf{Method} & \textbf{Mod.} & \textbf{Task} & \textbf{NDS}$\uparrow$ & \textbf{mAP}$\uparrow$ & \textbf{mIoU-7}$\uparrow$ & \textbf{mIoU}$^*$$\uparrow$ \\
\midrule
\multicolumn{7}{@{}l}{\textit{Single-task detection}} \\
CenterFusion$^*$~\cite{nabati2021centerfusion} & R+C & Det & 45.3 & 33.2 & -- & -- \\
CRAFT$^*$~\cite{kim2023craft} & R+C & Det & 51.7 & 41.1 & -- & -- \\
CRN$^*$~\cite{crn} (R50) & R+C & Det & 56.0 & 49.0 & -- & -- \\
RCBEVDet$^*$~\cite{rcbevdet} (R50) & R+C & Det & 56.8 & 45.3 & -- & -- \\
HyDRa$^*$~\cite{wolters2025unleashing} (R50) & R+C & Det & 58.5 & 49.4 & -- & -- \\
CRT-Fusion$^*$~\cite{kim2024crt} (R50) & R+C & Det & 59.7 & 50.8 & -- & -- \\
RCBEVDet++$^*$~\cite{lin2024rcbevdet++} (R50) & R+C & Det & 60.4 & 51.9 & -- & -- \\
RaCFormer$^*$~\cite{chu2025racformer} (R50) & R+C & Det & \textbf{61.3} & \textbf{54.1} & -- & -- \\
\midrule
\multicolumn{7}{@{}l}{\textit{Single-task segmentation}} \\
BEVGuide$^*$~\cite{Man_2023_CVPR} (EffNet) & R+C & Seg & -- & -- & -- & 60.0$^b$ \\
RCBEVDet++$^*$~\cite{lin2024rcbevdet++} (R101) & R+C & Seg & -- & -- & -- & 62.8$^b$ \\
BEVCar$^*$~\cite{bevcar} (ViT-B) & R+C & Seg & -- & -- & -- & \textbf{70.9}$^a$ \\
RESAR-BEV$^*$~\cite{zeng2026resar} (R101) & R+C & Seg & -- & -- & \textbf{54.0}$^c$ & -- \\
\midrule
\multicolumn{7}{@{}l}{\textit{R+C joint det+seg}} \\
(A) MTL baseline (R50) & R+C & Det+Seg & \textbf{55.7} & \textbf{44.4} & 42.9 & 50.4 \\
(B) MTL + CTAB$^\ddagger$ (R50) & R+C & Det+Seg & 55.6 & \textbf{44.4} & \textbf{43.5} & \textbf{51.0} \\
\bottomrule
\end{tabular}%
}
\end{table}

Table~\ref{tab:comparison} positions our work against existing radar-camera methods.
All prior R+C methods address detection or segmentation independently; ours is the first to jointly optimize both with explicit cross-task interaction.
Our MTL+CTAB model achieves 55.6~NDS and 43.5~mIoU-7 simultaneously, reaching 51.0~mIoU-4 on our 4-class subset, with 44.4~mAP.
While camera-only multi-task methods (BEVerse~\cite{beverse}, HENet~\cite{xia2024henet}) and LiDAR-camera methods (MaskBEV~\cite{maskbev}) exist, none operate in the radar-camera setting.
Our work brings cross-task attention to radar-camera fusion at a fraction of the sensor cost.

Table~\ref{tab:main_results} isolates the CTAB contribution.
Compared to the MTL baseline without cross-task interaction, CTAB improves mIoU-7 by $+0.6$ points and mIoU-4 by $+0.6$ points at essentially neutral detection ($-0.1$ NDS, $0.0$ mAP).
This indicates that bidirectional feature exchange benefits segmentation substantially without interfering with detection: segmentation gains object-level geometric cues from detection, and the detector retains its full accuracy.
Notably, CTAB achieves this with only 0.58M additional parameters---less than half the decoder (1.33M)---demonstrating that the cross-task signal is highly parameter-efficient.

Per-class results (Table~\ref{tab:seg_perclass}) reveal that CTAB's gains are not uniform: they concentrate on the thin classes where the baseline is weakest---pedestrian crossing ($+1.8$), stop line ($+1.8$), and divider ($+0.7$)---whose small spatial footprint ($\sim$15--30~cm) and sparse radar returns benefit from the object-level cues injected by the detection branch. Drivable area also benefits ($+0.6$), while the remaining dense classes (carpark, walkway, vehicle) move by at most $0.2$~pp, consistent with the baseline already capturing most of their structure. This confirms that cross-task attention helps most where one task's signal is intrinsically weak.

Camera-only multi-task methods demonstrate joint perception but lack radar's depth and velocity robustness.
In the LiDAR-camera domain, MaskBEV~\cite{maskbev} and M\textsuperscript{3}Net~\cite{chen2025m3net} achieve stronger absolute numbers, but rely on dense LiDAR point clouds unavailable in cost-sensitive deployments.
Our work brings cross-task attention to the radar-camera setting for the first time, bridging this modality gap.

\begin{table}[t]
\centering
\caption{Multi-task ablation on nuScenes val. All models use our RCBEVDet-based MTL framework with ResNet-50, SegDecoderV2 (IN, 3 blocks), and BEV Upsampling to $200 \times 200$.}
\label{tab:main_results}
\renewcommand{\arraystretch}{1.15}
\resizebox{\columnwidth}{!}{%
\begin{tabular}{@{}l cccc c@{}}
\toprule
\textbf{Configuration} & \textbf{NDS}$\uparrow$ & \textbf{mAP}$\uparrow$ & \textbf{mIoU-7}$\uparrow$ & \textbf{mIoU-4}$\uparrow$ & \textbf{+Params} \\
\midrule
RCBEVDet$^*$ (det only) & 56.8 & 45.3 & -- & -- & -- \\
\midrule
(A) MTL baseline & \textbf{55.7} & \textbf{44.4} & 42.9 & 50.4 & +1.63M \\
(B) MTL + CTAB & 55.6 & \textbf{44.4} & \textbf{43.5} & \textbf{51.0} & +2.21M \\
\midrule
\multicolumn{6}{@{}l}{\textit{CTAB improvement}} \\
$\Delta$ (B$-$A) & $-0.1$ & $0.0$ & \textbf{+0.6} & \textbf{+0.6} & +0.58M \\
\bottomrule
\end{tabular}%
}
\end{table}

\textbf{Efficiency.}
The multi-task extension adds 2.21M parameters (decoder 1.33M + upsampling 0.30M + CTAB 0.58M) over the 38.4M detection backbone, a 5.8\% increase.
CTAB itself represents only 1.4\% additional parameters.
The multi-task model processes both tasks in a single forward pass, avoiding the ${\sim}2\times$ backbone cost of running separate detection and segmentation pipelines.

\subsection{Gate Analysis}

\begin{figure}[t]
\centering
\includegraphics[width=\columnwidth]{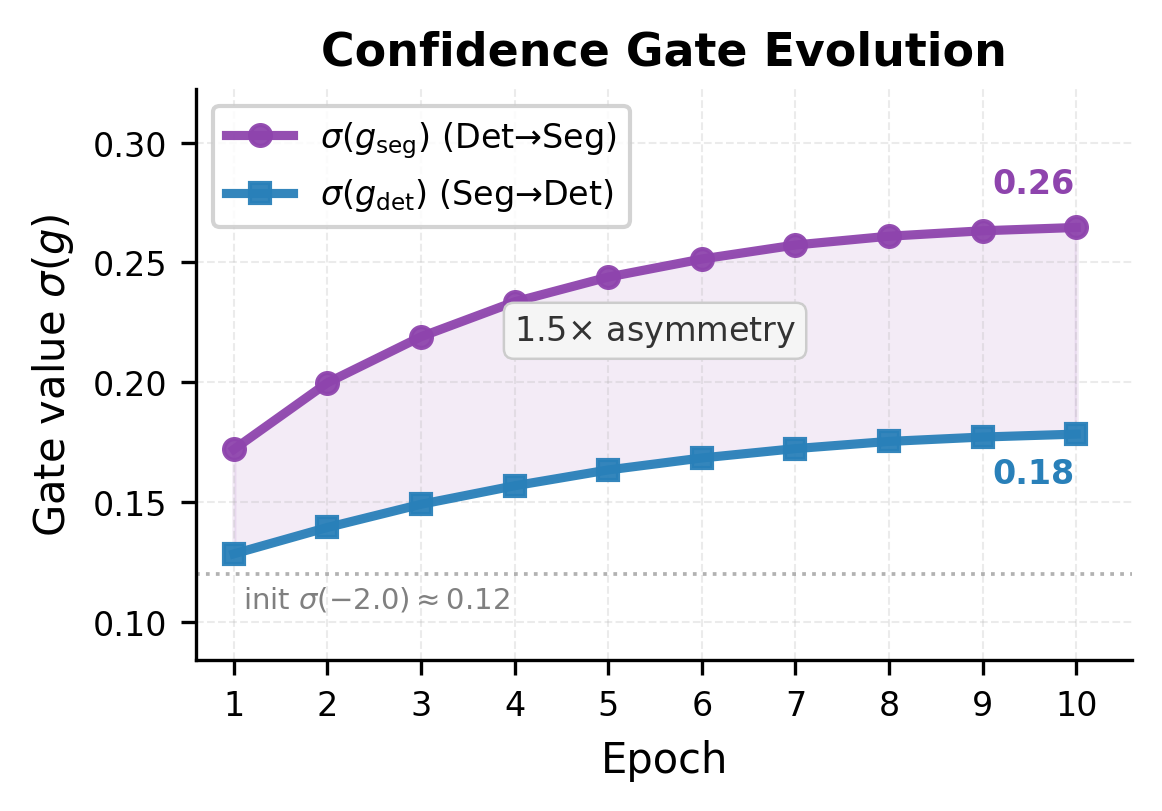}
\caption{\textbf{Confidence gate evolution during training (Exp~B).}
Both gates are initialized at $\sigma(-2.0) \approx 0.12$.
The segmentation gate $\sigma(g_{\text{seg}})$ (purple) rises faster than the detection gate $\sigma(g_{\text{det}})$ (blue), indicating that the segmentation branch benefits more from cross-task features.
This asymmetry emerges entirely from learning---both gates share identical architecture and initialization.
}
\label{fig:gate_evolution}
\end{figure}

Fig.~\ref{fig:gate_evolution} visualizes the evolution of CTAB's confidence gates during training.
Both gates remain well below unity, indicating that CTAB acts as a gentle residual refinement rather than a dominant signal---consistent with the small, steady improvement over the baseline.
The segmentation gate rises more markedly (${\approx}0.17 \to 0.26$) than the detection gate (${\approx}0.13 \to 0.18$), yielding a ${\approx}1.5\times$ asymmetry that emerges from learning: detection features supply object-level priors that sharpen segmentation boundaries, while the pretrained detector---already anchored to the full BEV---needs less help from segmentation context.
This consistent asymmetry validates using learnable per-branch gates over a fixed weighting.

\subsection{Segmentation Per-Class Results}

\begin{table}[t]
\centering
\caption{Per-class BEV segmentation IoU (\%) on nuScenes val at $200 \times 200$ resolution. IoU@max reported. Divider combines road and lane dividers.}
\label{tab:seg_perclass}
\renewcommand{\arraystretch}{1.15}
\resizebox{\columnwidth}{!}{%
\begin{tabular}{@{}l ccccccc c@{}}
\toprule
\textbf{Method} & \rotatebox{70}{Drivable} & \rotatebox{70}{Carpark} & \rotatebox{70}{Ped. Cross.} & \rotatebox{70}{Walkway} & \rotatebox{70}{Stop Line} & \rotatebox{70}{Divider} & \rotatebox{70}{Vehicle} & \textbf{mIoU-7} \\
\midrule
(A) Baseline & 72.7 & \textbf{38.6} & 34.7 & \textbf{44.5} & 26.1 & 33.7 & 49.8 & 42.9 \\
(B) + CTAB & \textbf{73.3} & 38.4 & \textbf{36.5} & 44.4 & \textbf{27.9} & \textbf{34.4} & 49.8 & \textbf{43.5} \\
\midrule
$\Delta$ & \textbf{+0.6} & $-0.2$ & \textbf{+1.8} & $-0.1$ & \textbf{+1.8} & \textbf{+0.7} & $0.0$ & \textbf{+0.6} \\
\bottomrule
\end{tabular}%
}
\end{table}

The $200 \times 200$ resolution ($0.5\,\text{m/cell}$) enables finer delineation of thin structures such as dividers and stop lines compared to the $128 \times 128$ backbone grid ($0.8\,\text{m/cell}$).
The addition of the vehicle class, derived from 3D bounding box annotations projected to BEV, provides a compact evaluation subset; however, we note that prior methods report mIoU over different class sets (Table~\ref{tab:comparison}), so cross-method comparison should be interpreted with caution.

\begin{figure}[t]
\centering
\includegraphics[width=\columnwidth]{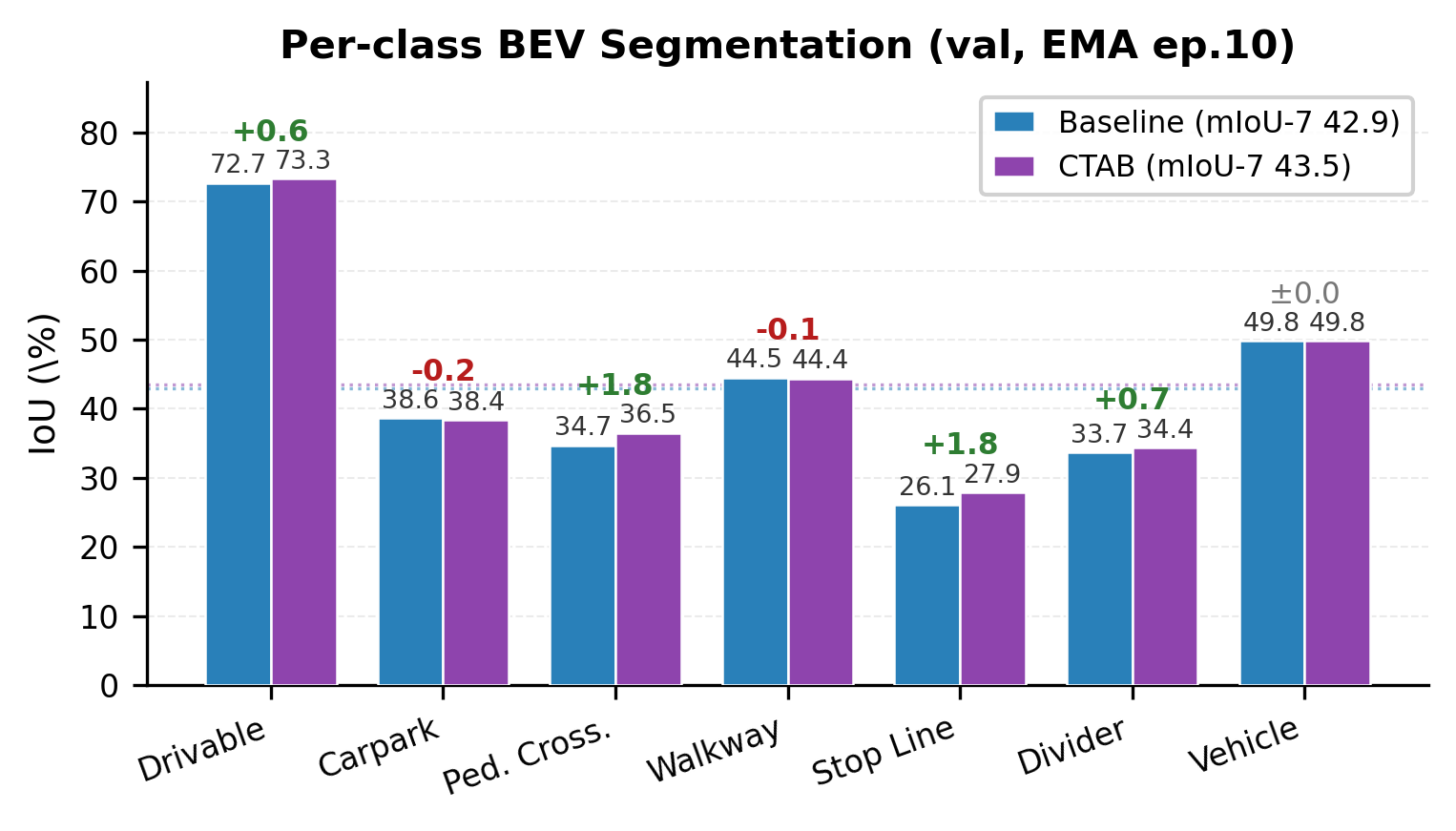}
\caption{\textbf{Per-class BEV segmentation IoU on nuScenes val.}
Grouped bars compare Baseline (blue) and CTAB (purple) across the seven map classes; dotted lines mark the overall mIoU-7.
CTAB's gains concentrate on thin and sparse classes---pedestrian crossing ($+1.8$), stop line ($+1.8$), and divider ($+0.7$)---while dense classes move by at most $0.2$~pp in either direction.
Vehicle IoU is unchanged at 49.8 in both models, reflecting that 3D box-derived GT masks are already well-captured by detection-branch features.
}
\label{fig:perclass}
\end{figure}

\subsection{Qualitative Results}

\begin{figure*}[t]
\centering
\includegraphics[width=\textwidth]{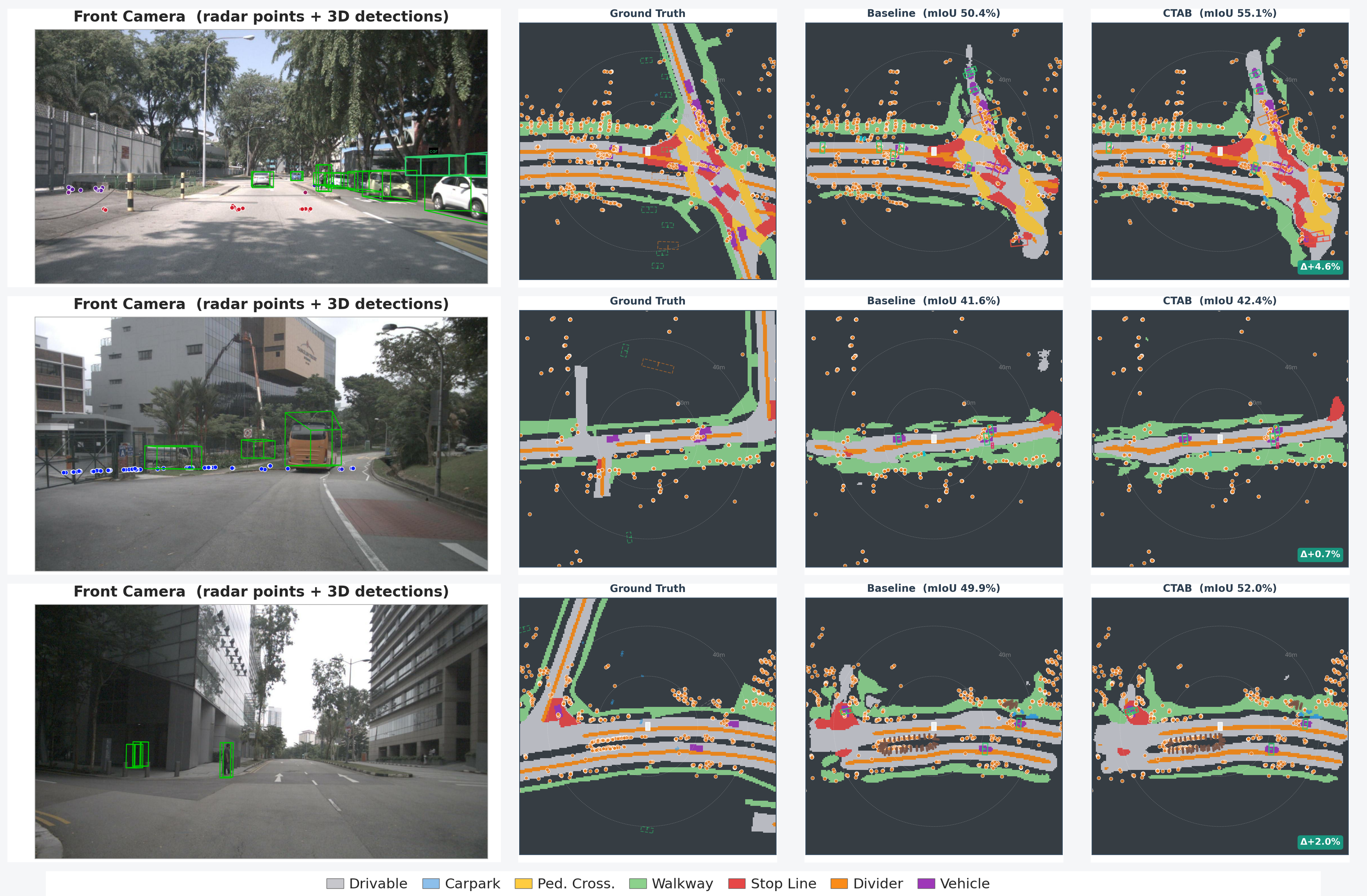}
\caption{\textbf{Qualitative comparison on three nuScenes validation scenes.}
Each row shows a different scene.
\textit{Column~1}: Front camera with projected radar points and 3D detection boxes.
\textit{Column~2}: Ground Truth BEV segmentation.
\textit{Columns~3--4}: Baseline and CTAB BEV segmentation with detected 3D boxes (colored by class) and radar point cloud (orange dots).
Per-scene mIoU and the CTAB improvement ($\Delta$) are shown.
}
\label{fig:qualitative}
\end{figure*}

Fig.~\ref{fig:qualitative} visualizes the full multi-task output across three diverse scenes, combining BEV segmentation, 3D detection boxes, and radar point cloud overlay.
CTAB produces tighter vehicle segmentation masks that better align with detected 3D bounding boxes, while radar points (orange) cluster on detected vehicles, confirming the spatial coherence of our radar-camera framework.
Lane and road dividers appear more continuous in CTAB predictions, and transitions between drivable area and walkway are sharper.

Fig.~\ref{fig:perclass} further isolates per-class segmentation quality, showing IoU gains for both large-area classes (drivable area) and thin structures (lane dividers).

\subsection{Discussion}

Our results demonstrate that cross-task attention is a viable and effective mechanism for radar-camera BEV multi-task learning.
Several observations merit discussion.

\textbf{Multi-task optimization.}
A common concern in multi-task learning is negative transfer, where jointly training tasks degrades individual task performance compared to single-task training.
Our baseline model (A) achieves 55.7~NDS compared to the single-task RCBEVDet~\cite{rcbevdet}'s 56.8 NDS, indicating a modest detection degradation from shared training.
CTAB largely preserves this level ($55.6$ NDS), suggesting that cross-task interaction can at least neutralize the negative transfer typically observed in multi-task BEV perception.

\textbf{Normalization and training stability.}
The choice of Instance Normalization for the decoder and Group Normalization for CTAB proved essential for stable training.
In early experiments with Batch Normalization, we observed significant train-eval gaps: the segmentation branch performed well during training but degraded substantially at evaluation time, when BN switches from batch to running statistics.
This effect is amplified by the multi-view BEV setting, where each image contributes a distinct viewpoint, reducing effective batch diversity; IN and GN eliminate this by normalizing within each sample.

\textbf{Resolution asymmetry.}
Operating detection at $128 \times 128$ and segmentation at $200 \times 200$ reflects the different spatial requirements of each task.
Detection regression targets (centroids, sizes) are relatively robust to discretization, while segmentation of thin structures requires finer spatial resolution.
BEV Upsampling enables this asymmetry without modifying the shared backbone, preserving the pretrained detection head while improving segmentation.

\section{Conclusion}
\label{sec:conclusion}

We presented CTAB, a lightweight bidirectional cross-task attention module for joint radar-camera BEV detection and segmentation.
By leveraging multi-scale deformable attention with confidence gating and group normalization, CTAB enables stable and effective feature exchange between detection and segmentation branches, adding only 0.58M parameters.
Our framework incorporates an enhanced segmentation decoder with Instance Normalization and learnable BEV Upsampling from $128 \times 128$ to $200 \times 200$ resolution.
Integrated into a radar-camera BEV multi-task framework with HUW-based loss balancing, CTAB improves segmentation over the multi-task baseline while maintaining competitive detection performance, with learned confidence gates revealing meaningful asymmetry that validates the complementary nature of cross-task information flow.

\textbf{Limitations.}
While BEV Upsampling improves segmentation resolution, the underlying backbone features remain at $0.8\,\text{m/cell}$; the upsampling module refines spatial detail but cannot recover information lost during the initial BEV projection.
Our ResNet-50 backbone is smaller than the ViT-B backbones used by top segmentation methods~\cite{bevcar}, limiting direct mIoU comparison; however, the within-framework ablation between baseline and CTAB remains valid.

\textbf{Future directions.}
Replacing ResNet-50 with self-supervised vision transformers should narrow the gap to ViT-based segmentation, particularly for thin structures.
CTAB's scalar confidence gates could be replaced by per-location uncertainty maps $U(x,y)$, modulating the cross-task signal selectively---strong near objects, suppressed in empty regions---potentially benefiting dense classes.
Task-adaptive radar fusion---weighting radar features differently per task rather than sharing a single BEV---is also promising, given the different sensitivity of detection and segmentation to radar signals.


\bibliographystyle{IEEEtran}
\bibliography{references}

@article{bevdet,
  title={{BEVDet: High-Performance Multi-Camera 3D Object Detection in Bird-Eye-View}},
  author={Huang, Junjie and Huang, Guan and Zhu, Zheng and Ye, Yun and Du, Dalong},
  journal={arXiv preprint arXiv:2112.11790},
  year={2021}
}

@inproceedings{bevdepth,
  title={{BEVDepth: Acquisition of Reliable Depth for Multi-View 3D Object Detection}},
  author={Li, Yinhao and Ge, Zheng and Yu, Guanyi and Yang, Jinrong and Wang, Zengran and Shi, Yukang and Sun, Jianjian and Li, Zeming},
  booktitle={{Proceedings of the AAAI Conference on Artificial Intelligence}},
  volume={37},
  number={2},
  pages={1477--1485},
  year={2023}
}

@article{beverse,
  title={{BEVerse: Unified Perception and Prediction in Birds-Eye-View for Vision-Centric Autonomous Driving}},
  author={Zhang, Yunpeng and Zhu, Zheng and Zheng, Wenzhao and Huang, Junjie and Huang, Guan and Zhou, Jie and Lu, Jiwen},
  journal={arXiv preprint arXiv:2205.09743},
  year={2022}
}

@inproceedings{crn,
  title={{CRN: Camera Radar Net for Accurate, Robust, Efficient 3D Perception}},
  author={Kim, Youngseok and Shin, Juyeb and Kim, Sanmin and Lee, In-Jae and Choi, Jun Won and Kum, Dongsuk},
  booktitle={{Proceedings of the IEEE/CVF International Conference on Computer Vision}},
  pages={17615--17626},
  year={2023}
}

@inproceedings{rcbevdet,
  title={{RCBEVDet: Radar-Camera Fusion in Bird's Eye View for 3D Object Detection}},
  author={Lin, Zhiwei and Liu, Zhe and Xia, Zhongyu and Wang, Xinhao and Wang, Yongtao and Qi, Shengxiang and Dong, Yang and Dong, Nan and Zhang, Le and Zhu, Ce},
  booktitle={{Proceedings of the IEEE/CVF Conference on Computer Vision and Pattern Recognition}},
  pages={14928--14937},
  year={2024}
}

@inproceedings{bevcar,
  title={{BEVCar: Camera-Radar Fusion for BEV Map and Object Segmentation}},
  author={Schramm, Jonas and V{\"o}disch, Niclas and Petek, K{\"u}rsat and Kiran, B Ravi and Yogamani, Senthil and Burgard, Wolfram and Valada, Abhinav},
  booktitle={{2024 IEEE/RSJ International Conference on Intelligent Robots and Systems (IROS)}},
  pages={1435--1442},
  year={2024},
  organization={IEEE}
}

@article{milli2023multi,
  title={{Multi-Modal Multi-Task (3MT) Road Segmentation}},
  author={Milli, Erkan and Erkent, {\"O}zg{\"u}r and Y{\i}lmaz, As{\i}m Egemen},
  journal={IEEE Robotics and Automation Letters},
  volume={8},
  number={9},
  pages={5408--5415},
  year={2023},
  publisher={IEEE}
}

@inproceedings{uniad,
  title={{Planning-Oriented Autonomous Driving}},
  author={Hu, Yihan and Yang, Jiazhi and Chen, Li and Li, Keyu and Sima, Chonghao and Zhu, Xizhou and Chai, Siqi and Du, Senyao and Lin, Tianwei and Wang, Wenhai and others},
  booktitle={{Proceedings of the IEEE/CVF Conference on Computer Vision and Pattern Recognition}},
  pages={17853--17862},
  year={2023}
}

@inproceedings{bevfusion,
  title={{BEVFusion: Multi-Task Multi-Sensor Fusion with Unified Bird's-Eye View Representation}},
  author={Liu, Zhijian and Tang, Haotian and Amini, Alexander and Yang, Xinyu and Mao, Huizi and Rus, Daniela L and Han, Song},
  booktitle={{2023 IEEE International Conference on Robotics and Automation (ICRA)}},
  pages={2774--2781},
  year={2023},
  organization={IEEE}
}

@inproceedings{maskbev,
  title={{MaskBEV: Towards a Unified Framework for BEV Detection and Map Segmentation}},
  author={Zhao, Xiao and Zhang, Xukun and Yang, Dingkang and Sun, Mingyang and Li, Mingcheng and Wang, Shunli and Zhang, Lihua},
  booktitle={{Proceedings of the 32nd ACM International Conference on Multimedia}},
  pages={2652--2661},
  year={2024}
}

@inproceedings{padnet,
  title={{PAD-Net: Multi-Tasks Guided Prediction-and-Distillation Network for Simultaneous Depth Estimation and Scene Parsing}},
  author={Xu, Dan and Ouyang, Wanli and Wang, Xiaogang and Sebe, Nicu},
  booktitle={{Proceedings of the IEEE Conference on Computer Vision and Pattern Recognition}},
  pages={675--684},
  year={2018}
}

@inproceedings{mtinet,
  title={{MTI-Net: Multi-Scale Task Interaction Networks for Multi-Task Learning}},
  author={Vandenhende, Simon and Georgoulis, Stamatios and Van Gool, Luc},
  booktitle={{European Conference on Computer Vision}},
  pages={527--543},
  year={2020},
  organization={Springer}
}

@inproceedings{xu2022mtformer,
  title={{MTFormer: Multi-Task Learning via Transformer and Cross-Task Reasoning}},
  author={Xu, Xiaogang and Zhao, Hengshuang and Vineet, Vibhav and Lim, Ser-Nam and Torralba, Antonio},
  booktitle={{European Conference on Computer Vision}},
  pages={304--321},
  year={2022},
  organization={Springer}
}

@inproceedings{zhudeformable,
  title={{Deformable DETR: Deformable Transformers for End-to-End Object Detection}},
  author={Zhu, Xizhou and Su, Weijie and Lu, Lewei and Li, Bin and Wang, Xiaogang and Dai, Jifeng},
  booktitle={{International Conference on Learning Representations}},
  year={2021}
}

@inproceedings{wolters2025unleashing,
  title={{Unleashing HyDRa: Hybrid Fusion, Depth Consistency and Radar for Unified 3D Perception}},
  author={Wolters, Philipp and Gilg, Johannes and Teepe, Torben and Herzog, Fabian and Laouichi, Anouar and Hofmann, Martin and Rigoll, Gerhard},
  booktitle={{2025 IEEE International Conference on Robotics and Automation (ICRA)}},
  pages={7467--7474},
  year={2025},
  organization={IEEE}
}

@article{kim2024crt,
  title={{CRT-Fusion: Camera, Radar, Temporal Fusion Using Motion Information for 3D Object Detection}},
  author={Kim, Jisong and Seong, Minjae and Choi, Jun Won},
  journal={Advances in Neural Information Processing Systems},
  volume={37},
  pages={108625--108648},
  year={2024}
}

@article{lin2024rcbevdet++,
  title={{RCBEVDet++: Toward High-Accuracy Radar-Camera Fusion 3D Perception Network}},
  author={Lin, Zhiwei and Liu, Zhe and Wang, Yongtao and Zhang, Le and Zhu, Ce},
  journal={arXiv preprint arXiv:2409.04979},
  year={2024}
}

@inproceedings{chu2025racformer,
  title={{RaCFormer: Towards High-Quality 3D Object Detection via Query-Based Radar-Camera Fusion}},
  author={Chu, Xiaomeng and Deng, Jiajun and You, Guoliang and Duan, Yifan and Li, Houqiang and Zhang, Yanyong},
  booktitle={{Proceedings of the IEEE/CVF Conference on Computer Vision and Pattern Recognition}},
  pages={17081--17091},
  year={2025}
}

@inproceedings{harley2023simple,
  title={{Simple-BEV: What Really Matters for Multi-Sensor BEV Perception?}},
  author={Harley, Adam W and Fang, Zhaoyuan and Li, Jie and Ambrus, Rares and Fragkiadaki, Katerina},
  booktitle={{2023 IEEE International Conference on Robotics and Automation (ICRA)}},
  pages={2759--2765},
  year={2023},
  organization={IEEE}
}

@inproceedings{Man_2023_CVPR,
  title={{BEV-Guided Multi-Modality Fusion for Driving Perception}},
  author={Man, Yunze and Gui, Liang-Yan and Wang, Yu-Xiong},
  booktitle={{Proceedings of the IEEE/CVF Conference on Computer Vision and Pattern Recognition (CVPR)}},
  month={June},
  year={2023},
  pages={21960--21969}
}

@article{zeng2026resar,
  title={{RESAR-BEV: An Explainable Progressive Residual Autoregressive Approach for Camera-Radar Fusion in BEV Segmentation}},
  author={Zeng, Zhiwen and Yin, Yunfei and Yuan, Zheng and Dey, Argho and Bao, Xianjian},
  journal={IEEE Transactions on Intelligent Transportation Systems},
  year={2026},
  publisher={IEEE}
}

@inproceedings{xia2024henet,
  title={{HENet: Hybrid Encoding for End-to-End Multi-Task 3D Perception from Multi-View Cameras}},
  author={Xia, Zhongyu and Lin, Zhiwei and Wang, Xinhao and Wang, Yongtao and Xing, Yun and Qi, Shengxiang and Dong, Nan and Yang, Ming-Hsuan},
  booktitle={{European Conference on Computer Vision}},
  pages={376--392},
  year={2024},
  organization={Springer}
}

@inproceedings{chen2025m3net,
  title={{M3Net: Multimodal Multi-Task Learning for 3D Detection, Segmentation, and Occupancy Prediction in Autonomous Driving}},
  author={Chen, Xuesong and Shi, Shaoshuai and Ma, Tao and Zhou, Jingqiu and See, Simon and Cheung, Ka Chun and Li, Hongsheng},
  booktitle={{Proceedings of the AAAI Conference on Artificial Intelligence}},
  volume={39},
  number={2},
  pages={2275--2283},
  year={2025}
}

@inproceedings{ye2023taskprompter,
  title={{TaskPrompter: Spatial-Channel Multi-Task Prompting for Dense Scene Understanding}},
  author={Ye, Hanrong and Xu, Dan},
  booktitle={{The Eleventh International Conference on Learning Representations}},
  year={2023}
}

@inproceedings{kendall2018multi,
  title={{Multi-Task Learning Using Uncertainty to Weigh Losses for Scene Geometry and Semantics}},
  author={Kendall, Alex and Gal, Yarin and Cipolla, Roberto},
  booktitle={{Proceedings of the IEEE Conference on Computer Vision and Pattern Recognition}},
  pages={7482--7491},
  year={2018}
}

@inproceedings{chen2018gradnorm,
  title={{GradNorm: Gradient Normalization for Adaptive Loss Balancing in Deep Multitask Networks}},
  author={Chen, Zhao and Badrinarayanan, Vijay and Lee, Chen-Yu and Rabinovich, Andrew},
  booktitle={{International Conference on Machine Learning}},
  pages={794--803},
  year={2018},
  organization={PMLR}
}

@article{yu2020gradient,
  title={{Gradient Surgery for Multi-Task Learning}},
  author={Yu, Tianhe and Kumar, Saurabh and Gupta, Abhishek and Levine, Sergey and Hausman, Karol and Finn, Chelsea},
  journal={Advances in Neural Information Processing Systems},
  volume={33},
  pages={5824--5836},
  year={2020}
}

@inproceedings{yin2021center,
  title={{Center-Based 3D Object Detection and Tracking}},
  author={Yin, Tianwei and Zhou, Xingyi and Krahenbuhl, Philipp},
  booktitle={{Proceedings of the IEEE/CVF Conference on Computer Vision and Pattern Recognition}},
  pages={11784--11793},
  year={2021}
}

@inproceedings{wu2018group,
  title={{Group Normalization}},
  author={Wu, Yuxin and He, Kaiming},
  booktitle={{Proceedings of the European Conference on Computer Vision (ECCV)}},
  pages={3--19},
  year={2018}
}

@inproceedings{caesar2020nuscenes,
  title={{nuScenes: A Multimodal Dataset for Autonomous Driving}},
  author={Caesar, Holger and Bankiti, Varun and Lang, Alex H and Vora, Sourabh and Liong, Venice Erin and Xu, Qiang and Krishnan, Anush and Pan, Yu and Baldan, Giancarlo and Beijbom, Oscar},
  booktitle={{Proceedings of the IEEE/CVF Conference on Computer Vision and Pattern Recognition}},
  pages={11621--11631},
  year={2020}
}

@misc{contributors2020mmdetection3d,
  title={{MMDetection3D: OpenMMLab Next-Generation Platform for General 3D Object Detection}},
  author={Contributors, MMDetection3D},
  year={2020}
}

@inproceedings{nabati2021centerfusion,
  title={{CenterFusion: Center-Based Radar and Camera Fusion for 3D Object Detection}},
  author={Nabati, Ramin and Qi, Hairong},
  booktitle={{Proceedings of the IEEE/CVF Winter Conference on Applications of Computer Vision}},
  pages={1527--1536},
  year={2021}
}

@inproceedings{kim2023craft,
  title={{CRAFT: Camera-Radar 3D Object Detection with Spatio-Contextual Fusion Transformer}},
  author={Kim, Youngseok and Kim, Sanmin and Choi, Jun Won and Kum, Dongsuk},
  booktitle={{Proceedings of the AAAI Conference on Artificial Intelligence}},
  volume={37},
  number={1},
  pages={1160--1168},
  year={2023}
}

@inproceedings{huang2023fuller,
  title={{FULLER: Unified Multi-Modality Multi-Task 3D Perception via Multi-Level Gradient Calibration}},
  author={Huang, Zhijian and Lin, Sihao and Liu, Guiyu and Luo, Mukun and Ye, Chaoqiang and Xu, Hang and Chang, Xiaojun and Liang, Xiaodan},
  booktitle={{Proceedings of the IEEE/CVF International Conference on Computer Vision}},
  pages={3502--3511},
  year={2023}
}

@inproceedings{ge2023metabev,
  title={{MetaBEV: Solving Sensor Failures for 3D Detection and Map Segmentation}},
  author={Ge, Chongjian and Chen, Junsong and Xie, Enze and Wang, Zhongdao and Hong, Lanqing and Lu, Huchuan and Li, Zhenguo and Luo, Ping},
  booktitle={{Proceedings of the IEEE/CVF International Conference on Computer Vision}},
  pages={8721--8731},
  year={2023}
}

\end{document}